\documentclass[lettersize,journal]{IEEEtran}

\usepackage[ruled,vlined]{algorithm2e}
\usepackage{graphicx}

\usepackage{graphicx}
\usepackage{amsmath, amssymb}
\usepackage{booktabs}
\usepackage{xcolor}

\begin{document}

\title{GUESS: Generative Uncertainty Ensemble for Self Supervision}

\author{Salman Mohamadi, Gianfranco Doretto, Donald Adjeroh\\
  \textit{West Virginia University, Morgantown, USA}}




\maketitle

\begin{abstract}
   Self-supervised learning (SSL) frameworks consist of pretext task, and loss function aiming to learn useful general features from unlabeled data. 
The basic idea of most SSL baselines revolves around enforcing the invariance to a variety of data augmentations via the loss function. However, one main issue is that, 
inattentive or deterministic enforcement of the invariance to any kind of data augmentation is generally not only inefficient, but also potentially detrimental to performance on the downstream tasks. 
In this work, we investigate the issue from the viewpoint of uncertainty in invariance representation. Uncertainty representation is fairly under-explored in the design of SSL architectures as well as loss functions. 
We incorporate uncertainty representation in both loss function as well as architecture design aiming for more data-dependent invariance enforcement. The former 
is represented in the form of data-derived uncertainty in SSL loss function resulting in a \textbf{generative-discriminative loss function}. The latter is achieved 
by feeding slightly different distorted versions of samples to the ensemble aiming for learning better and more robust representation.
Specifically, building upon the 
recent methods  that use hard and soft whitening (a.k.a redundancy reduction), we introduce a new approach GUESS, a pseudo-whitening framework, composed of controlled uncertainty injection, a new architecture, and a new loss function.
We include detailed results and ablation analysis establishing GUESS as a new baseline.
\end{abstract}


 \section{Introduction}
\label{sec:intro}
\IEEEPARstart{S}{elf}-supervised learning (SSL) emerged as a framework to use unlabeled data in a supervised manner aiming for learning useful representation \cite{jing2020self,zhou2023unified}. SSL along with deep active learning \cite{mohamadi2022deep,mohamadi2020deep} and semi-supervised learning \cite{yang2022survey} are among label-efficient frameworks within deep learning. Essentially, with SSL, the supervision signal comes from the data itself, as opposed to human annotation effort. Interestingly, the performance of recent SSL frameworks have been shown to be quite competitive with, and at times superior to, those from supervised learning techniques on various downstream visual tasks \cite{mohamadi2024active,zbontar2021barlow}. 
SSL generally aims at learning the invariant representation of the data, through an evolving set of approaches \cite{mohamadi2023fussl,mohamadi2023more}. 
From a reductionist point of view, as a proxy task \cite{huang2022learning}, the model is to learn how to represent the augmented views of the input data in such a way that the views that come from the same sample (also called positive examples) have as similar a representation as possible, while differentiating them as much as possible from the representations for augmented views that come from other sample data (negative examples). A popular class of SSL techniques are contrastive methods \cite{oord2018representation,tian2020contrastive,he2020momentum,chen2020simple,bachman2019learning}, 
which primarily contrast positive 
examples against both positive and negative examples.
Another set of approaches, a.k.a non-constrastive approaches, such as \cite{grill2020bootstrap,chen2021exploring}, however, show that it is possible to produce similar or superior results to the contrastive approaches, without necessarily requiring to contrast against negative examples. Tian et al. \cite{tian2021understanding}  provided an analysis of this later set of approaches and showed how they actually avoid trivial solutions (representation collapse), a challenging problem for the contrastive learning techniques. Following this trend, i.e., negative-free techniques, two recent approaches \cite{ermolov2021whitening,zbontar2021barlow} emerged based on redundancy reduction representation learning. They considered the notion of whitening, namely hard whitening  \cite{ermolov2021whitening}, and soft whitening \cite{zbontar2021barlow}, respectively, which enjoy optimisation of conceptually similar mathematical constraints in a non-contrastive framework. Essentially, both methods at some point implicitly argue that, the cross-correlation between the embedding/latent space of a pair of symmetric networks trained on augmented views of the same sample should result in an identity matrix of proper size.
\\
\indent
One major
problem of all the mentioned work which recently gained attention is that aggressive/blind enforcement of the invariance to any kind of data augmentation is generally not only inefficient but potentially detrimental to downstream tasks \cite{huang2022learning}. Some methods 
implicitly addressed this relying on the general idea of incorporating a prior from the downstream task. These approaches include pixel consistency \cite{wang2021dense,xie2021propagate}, bounding-box consistency \cite{roh2021spatially,xiao2021region} and mask consistency \cite{oord2018representation} between instances. A recent work based on spatial alignment \cite{huang2022learning} also argue that former techniques due to using task-specific priors, are not generalizable to other tasks. However we investigate the issue from the perspective of SSL loss function and architecture aiming for modifying invariance enforcement. We closely observed this issue along with the evolution of SSL methods, and came up with a general SSL framework built on recent whitening based approaches developed based on redundancy reduction. Our approach navigates through the observation that blind invariance enforcement (as opposed to our data-dependent invariance enforcement)  at best, only ensures semantics similarity enforcement and disregard overall appearance \cite{fini2022self,huang2022learning} and potentially some useful information. The beauty of our proposed work is that, it not only sets a new baseline relying on a general SSL framework, but also incorporates uncertainty representation in SSL, which in many senses has gone unnoticed since the beginning \cite{Lexpodcast,zbontar2021barlow}.

In fact, there have been recent observations on the need 
to consider uncertainty representation within the framework of representation learning methods for improved robustness and reliability \cite{Lexpodcast,zbontar2021barlow}. Some recent work \cite{hendrycks2019using,poggi2020uncertainty,liu2019exploiting} have considered uncertainty in their approaches, however, these were more concerned with the impact of SSL on model uncertainty estimation, and on robustness improvement. 
In this work, we explicitly assess how \textbf{controlled uncertainty derived from the data} can provide effective data-dependent invariance enforcement to SSL framework, hence impact the performance of such a framework on latter downstream tasks. This results in a new architecture as well as a new loss function with both generative and discriminative components.   
We investigate the effect of each contribution using an ablation study. 
In summary, our key contributions in this work are the following:

\begin{itemize}
\setlength\itemsep{-0.1em}
    \item Developing a data-dependent invariance enforcement technique using generative uncertainty representation, resulting in a generative-discriminative loss function. 
    \item Introducing uncertainty representation to the architecture, in the form of an ensemble of blocks with {block-specific} data augmentation and training, as well as introducing a trick to implement a computationally efficient version of ensemble. 
    \item Performing extensive experiments with six benchmark datasets on three downstream tasks, setting a new baseline to show the robustness and effectiveness of the proposed framework.
\end{itemize}

\section{Related work}
SSL gained  huge attention due to its effectiveness in learning from unlabeled data and its usefulness across many tasks including classification, segmentation, image enhancement \cite{liang2022self}, etc. SSL frameworks typically consist of a pretext task as well as an objective function \cite{jing2020self,wang2023self}.
There are a variety of pretext tasks for image data \cite{ermolov2021whitening,jing2020self}. On the other hand, loss functions are relatively less diversified. Generally speaking, there are two categories of loss functions, namely, \textbf{generative} loss functions, such as autoencoder reconstruction loss or adversarial loss \cite{donahue2016adversarial} in the context of GAN's discriminator and generator; as well as \textbf{discriminative} loss functions, such as triplet loss, contrastive loss, and non-contrastive loss. Frameworks based on contrastive loss such as \cite{oord2018representation,tian2020contrastive,he2020momentum,chen2020simple,bachman2019learning} enforce invariance to data augmentation by contrasting positive examples against both positive and negative examples, with the main downside of requiring a large batch of negative pairs.
This motivated approaches to SSL that primarily use only positive pairs (e.g.,  
BYOL \cite{grill2020bootstrap} and later SimSiam \cite{chen2021exploring}), 
while avoiding the problem of representation collapse or trivial representations. 
Most recent approaches are based on enforcing invariance using redundancy reduction through whitening the embedding/latent space \cite{ermolov2021whitening, zbontar2021barlow}. Accordingly, hard \cite{ermolov2021whitening} and soft \cite{zbontar2021barlow} whitening for self supervision not only improved on former baselines in general, but also came with simpler pipelines and less computation. One general problem is that, all these baselines aggressively enforce the invariance to data augmentation in some way. We note that recent work such as \cite{huang2022learning,roh2021spatially,wang2021dense,xiao2021region,xie2021propagate} implicitly address the inefficiency and detriments caused by aggressive enforcement of invariance to augmentation towards downstream task, though these are not easily generalizable to other tasks \cite{huang2022learning}. In this work, we take a fundamentally different approach in the sense that we do not address the issue by either a downstream task prior, or via spatial constraints on learning perturbed instances. Rather, we devise  \textbf{a general SSL framework} composed of controlled uncertainty injection, new  architecture, and new loss function, without needing auxiliary information.
We provide extensive relevant details on clustering based baselines \cite{caron2020unsupervised,caron2018deep,caron2021emerging} and several important loss functions, including triplet loss \cite{sohn2016improved}, typical contrastive loss \cite{wang2015unsupervised,oord2018representation,tian2020contrastive,he2020momentum,chen2020simple,bachman2019learning}, and non-contrastive loss functions \cite{grill2020bootstrap,chen2021exploring,zbontar2021barlow,ermolov2021whitening} in supplementary materials.

\section{Generative uncertainty ensemble for SSL}
\label{sec3into}
Building upon most recent baselines \cite{zbontar2021barlow,grill2020bootstrap}, whitening baselines, we devise a general SSL framework based on pseudo-whitening, which provides both data-dependent invariance enforcement as well as robustness. We performed a series of simple experiments as the initial experiments on pseudo-whitening, which support the arguments that 1) rigorous whitening would not always provide the best performance, and 2) rather a pseudo-whitening process guided by data-derived uncertainty could result in more improvement. 
Specifically we present the results of the Barlow-Twins in three different setting. First Barlow-Twins with strict whitening constraint (off-diagonal elements set to zero), then the results of whitening with a relaxed constraint. Finally the results of regularizing the whitening by adding some noise, a predefined matrix with Gaussian distributed off-diagonal elements, which essentially inject uncertainty into invariance enforcement. The experimental detail is in section 4.

In this work, uncertainty representation is encoded in two joint stages, pseudo-whitening of latent space as well as architecture in terms of ensemble representation.
To this end, we replace the identity matrix in the whitening loss function \cite{zbontar2021barlow} with a more data dependent matrix, that introduces controlled uncertainty in the analysis. Ultimately, this modified loss function coupled with the ensemble representation, allows our method to learn more robust and improved features via data-dependent invariance enforcement.   
Below we present our approach -- generative uncertainty ensemble for self supervision (GUESS) and elaborate on how we inject controlled uncertainty into the self supervision in two separate stages. Our loss function combines the power of both discriminative and generative loss functions.
The generative component helps shape the data derived controlled uncertainty (using autoencoder) in order to expose the discriminative component to possible uncertainty in invariance enforcement.
\begin{figure*}
\label{Fig2}
  \centering
  \includegraphics[scale=.415]{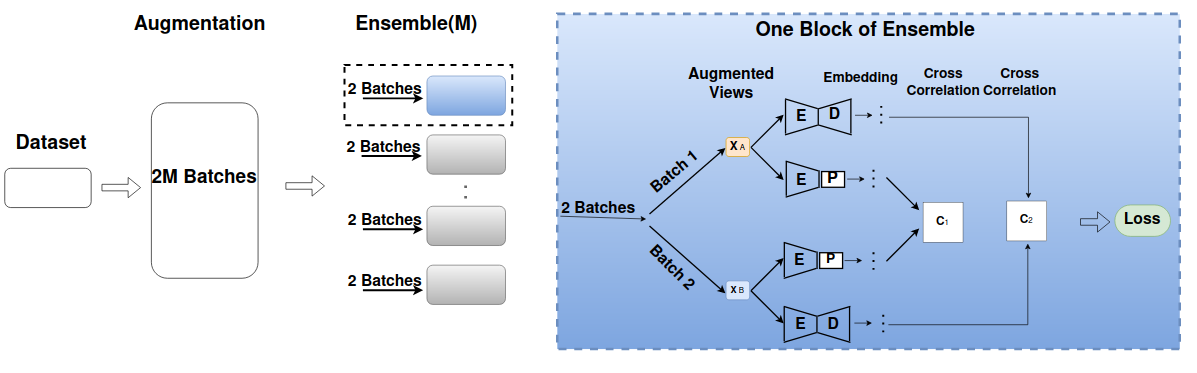}
		\caption{Schematic depiction of the proposed GUESS framework, with random allocation of augmented views to the blocks of ensemble-$M$. Each block gets its own set of augmented instances.} 
\end{figure*}

\subsection{Generative uncertainty block}
Hard whitening \cite{ermolov2021whitening} and soft whitening \cite{zbontar2021barlow} are both non-contrastive methods, and share some similarities, 
they learn an invariant representation on augmented instances of a sample, by performing whitening on the cross-correlation matrix of latent/embedding vectors.  We argue that encouraging zero off-diagonal elements in the cross-correlation matrix could however cause missing important discriminative information from the distorted views.

Rather than directly whitening the embedding space by forcing it to an identity matrix,
we aim for a \textbf{more data dependent constraint} to perform a pseudo-whitening of the embedding space. Thus, we introduce a so called generative uncertainty to the self supervision, to allow the model to consider both invariant and variant features captured by the diagonal and off-diagonal elements. Here the uncertainty comes from  enforcing the off-diagonal elements to values not necessarily zero, which are attained using  the cross-correlation of latent vectors of a pair of autoencoders (generative uncertainty). \textbf{In fact the loss function is no longer certain about setting off-diagonal directly to zero.}
In \cite{zbontar2021barlow}, the cross correlation of embedding vector of two identical architectures trained on two distorted views of a sample image are computed, and the optimization was mainly to minimize the difference between elements of the defined
cross-correlation matrix and the identity matrix.  
Consider a simple case, where two identical networks are fed with the same exact copy of a sample, as opposed to distorted views. Then reducing the problem to a deterministic one, it is more reasonable to expect the empirical cross-correlation matrix to be an identity matrix. Essentially, this ideal scenario corresponds to the case handled in \cite{zbontar2021barlow}, where the off-diagonal elements of cross-correlation are ignored as they are thought to be the cause of redundancy. Now, consider the more practical scenario where the networks are trained on augmented views of a set of samples. Ideally, the empirical cross-correlation matrix would not be an identity matrix, and would  depend on the type and level of distortion applied to the samples. Thus, a key question becomes how the framework can be modified to capture more robust features (both variant and invariant features), suitable for more realistic real-world scenarios, and possibly more robust to more distortions in the samples. To address this problem, we introduce two key innovations: 1) injection of controlled uncertainty via a data-dependent signal; 2) generation of this meaningful controlled signal using a modification of the basic architecture, adding a pair of new  autoencoders to the model, and introducing a second cross-correlation matrix. 

Let ${X_A}$ and ${X_B}$ denote  two sets of standard augmented versions generated for an  image. As shown in Fig. 1 (right),
the sets of images are fed separately to two identical networks, made from an encoder followed by a projector. The two are also fed separately to two identical autoencoders, where the embedding vectors of the autoencoders and the networks are of equal size. The cross-correlation between the normalized embedding of the networks are computed, and denoted  $C_1$. Further, we compute $C_2$, the cross-correlation between the normalized latent of each of the autoencoders. Unlike in \cite{zbontar2021barlow} which defined an objective function aimed at minimizing the distance between $C_1$ and ${I}$, an identity matrix of proper size, here the loss function is no longer certain about performing strict whitening, but rather tends to perform pseudo-whitening by defining a new matrix $C$ with entries between -1 and 1.
Matrix $C$ is essentially similar to ${I}$,  its diagonal entries are 1. However, its off-diagonal entries are non-zero, and are obtained from $C_2$, as related to the autoencoders. Obtaining the off-diagonal elements of $C$ from the cross-correlation between embedding of the autoencoders, allows injection of uncertainty into the strict whitening, where the whitening process is no longer certain about rigorously whitening the cross-correlation between embedding of the encoders, $C_1$ by setting of diagonal elements to zero. Rather, in this way it is indirectly affected by the level of distortion/augmentation implemented on the views, as the views are also fed to the autoencoders as the source of uncertainty for pseudo-whitening.
It is noteworthy to mention that the semantics captured in the generative uncertainty mainly relate to the features that autoencoder learns for near perfect reconstruction.

\subsection{Loss Functions}
Apart from a new architecture, our approach distinguishes itself from former approaches by its loss function. To learn the underlying representation using the pseudo-whitening approach, we minimize the total loss function $\mathcal{L}_t$, defined as follows: $\mathcal{L}_t = \mathcal{L}_w+\alpha \mathcal{L}_r$, where  $\mathcal{L}_w$ denotes the loss corresponding to pseudo-whitening with the $C_1$, and $\mathcal{L}_r$ denotes the autoencoders reconstruction loss, and $\alpha$ is a weighting factor. Since we have two autoencoders in a single block, $\mathcal{L}_r$ consists of two reconstruction losses, $\mathcal{L}_{r_1}$ and $\mathcal{L}_{r_{2}}$, each specifically a $L_2$ norm between input and the output. With $\beta$ as a weighting factor, $\mathcal{L}_w$ and $\mathcal{L}_r$ are then defined  as follows:  
\begin{equation}
    \label{eq1}
    \footnotesize
     \quad \mathcal{L}_w \triangleq \sum_{i}(1-C_{ii})^2 + \beta\sum_{i}\sum_{j\neq i}(C_{ij}-C_{1,ij})^2 ; \quad \mathcal{L}_r =\mathcal{L}_{r_1} + \mathcal{L}_{r_{2}}
\end{equation}
where $C_{ij}$ are elements of the cross-correlation matrix computed from the output vectors of the projector heads, while $C_{1,ij}$ are the elements of the cross-correlation matrix computed from the latent vectors of the autoencoders, as more discussed in Section 4.2. Here the optimizer is uncertain about strictly setting off-diagonal entries of matrix $C$ to zero.
Algorithm 1 shows the PyTorch-style pseudocode for a single generative uncertainty block.

\definecolor{commentcolor}{RGB}{110,154,155}   
\newcommand{\PyComment}[1]{\ttfamily\textcolor{commentcolor}{\# #1}}  
\newcommand{\PyCode}[1]{\ttfamily\textcolor{black}{#1}} 

\begin{algorithm}[h]
\small
\SetAlgoLined
    \PyComment{M: Network} 
    \PyComment{E: Autoencoder's Encoder, similar notation for both encoders} \\
    \PyComment{A: Autoencoder,  similar notation for both autodencoders} \\
    \PyComment{T: Augmentation} 
    \PyComment{$\alpha$, $\beta$, and $p$: Manually Set Hyperparameters } \\
    \PyComment{D: Dimension of Embedding}
    \PyComment{N: Batch Size}\\
    \PyComment{Norm: Z-Score Normalization } \\
    \PyComment{mm: Matrix-Matrix Multiplication } \\
    \PyComment{eye: Identity Matrix}\\
    \PyComment{C: Zero Matrix of Size $D\times D$}\\
    \PyCode{for x in loader}: \PyComment{Loading a Batch} \\
    \Indp   
        \PyComment{Random Augmentation} \\
        \PyCode{$T(x,p)= x_a, x_b$} \\ 
        \PyComment{Network Embedding}\\
        \PyCode{ $z_a^{(M)}$,\: $z_b^{(M)}=M(x_a)$,\: $M(x_b)$}\\
        \PyComment{Normalization}\\
        \PyCode{ $z_a^{(M)}$,\: $z_b^{(M)}=$Norm$(z_a^{(M)})$,\: Norm$(z_b^{(M)})$}\\
        \PyComment{Autoencoders Embeddings and Outputs}\\
        \PyCode{ $z_a^{(E)}$,\: $z_b^{(E)}=E(x_a)$,\: $E(x_b)$\:;\:\: $\hat{x_a}$,\: $\hat{x_b}=A(x_a)$,\: $A(x_b)$}\\
        \PyComment{Normalization}\\
        \PyCode{$z_a^{(E)}$,\: $z_b^{(E)}=$Norm$(z_a^{(E)})$,\: Norm$(z_b^{(E)})$\:;\:\: $\hat{x_a}$,\: $\hat{x_b}=$Norm$(\hat{x_a})$, Norm$(\hat{x_b})$}\\
        \PyComment{Cross-Correlation Matrix }\\
        \PyCode{$C_1= mm(z_a^{(M)}.T$,\: $z_b^{(M)})/N$,\quad $C_2= mm(z_a^{(E)}.T$,\: $z_b^{(E)})/N$}\\
        \PyComment{Pseudo-Whitening Matrix }\\
        \PyCode{diagonal$(C_2).mul\_ (0)$}\\
        \PyCode{$C=\left( eye(D)+\beta C_2 \right)$}  \PyComment{ eye(D): Identity }\\
        \PyComment{Loss }\\
        \PyCode{$\mathcal{L}_r=(x_a-\hat{x_a}).pow(2).sum()+(x_b-\hat{x_b}).pow(2).sum()$}\\
        \PyCode{ loss$= (C_1-C).pow(2).sum() + \mathcal{L}_r$}\PyComment{  $\mathcal{L}_{total}$ }\\
        \PyComment{Optimization}\\
        \PyCode{ loss.backward()} \\
        \PyCode{optimizer.step()}\\
    \Indm 
\caption{Summarized PyTorch-style pseudocode for one block of the ensemble}
\label{algo:your-algo}
\end{algorithm}

\subsection{Reducing computational complexity}
As the ensemble representation brings some computational overhead, we propose a more efficient version of the ensemble which reduces the computational complexity by half. We perform the experiments under both settings for comparison purposes.

\textbf{Ensemble:} As shown in Fig. 1, an ensemble of blocks are assembled, where each block is trained on its own specific distorted views of the samples. Essentially we assemble a varying number of generative uncertainty blocks to form an ensemble of blocks. To maximize  exploration in the representation space for each sample, these blocks of the ensemble are each trained independently with a different set of augmented examples (block specific augmented views). 
The proper number of augmented instances (which will depend on the size of the ensemble) will be generated, and separately fed to the blocks. Thus the use of an ensemble allows us to extract more information by learning a range of slightly different representations.
\\
\indent 
We denote the resulting framework as \textbf{GUESS-$m$}, where $m$ is the size, or number of generative uncertainty blocks. Thus GUESS-1 
will be  the basic case with just one block.

\textbf{Efficient ensemble with auto-correlation:} Relying on a new formulation, we 
simplify the architecture 
through replacing a pair of networks by a single network, and accordingly reformulate the loss function 
by replacing the cross-correlation by auto-correlation. This reduces the computational complexity by half. As shown in Fig. 2, we simplify the architecture of one block by computing the auto-correlation of one network for further pseudo-whitening, rather than cross-correlation of two networks. The corresponding loss function for this more efficient ensemble is presented later in this section. We denote the resulting framework as \textbf{GUESS-$m$-E}. It is noteworthy to mention that GUESS-1-E has essentially similar computational complexity as Barlow-Twins.

By \textbf{efficient ensemble} we reduce the computational complexity by half via a simpler architectural design and loss function, substituting the cross-correlation with auto-correlation. Corresponding loss function is as follows:

\begin{equation}
    \label{eq1}
    \footnotesize
     \quad \mathcal{L}_{w'} \triangleq \sum_{i}(1-C'_{ii})^2 + \beta\sum_{i}\sum_{j\neq i}(C'_{ij}-C"_{1,ij})^2 ; \quad \mathcal{L}_r =\mathcal{L}_{r_1} + \mathcal{L}_{r_{2}}
\end{equation}
where we have:
\begin{equation}
    C'_{ij} \triangleq \frac{\sum_{m'} z_{m',i}z_{m',j}}{\sqrt{\sum_m' (z_{m',i})^2} \sqrt{\sum_m' (z_{m',j})^2}}
\end{equation}
 where $z$ is the normalized output of projector head for one view, $x_1$, and $m'$ is  the batch size (note that similar to the  original framework, here for each sample we fed two views to the network).  Finally the elements of the matrix $C"$, auto-correlation matrix, is also computed from the latent space of one autoencoder similar to the equation for the elements of $C'$. 

\begin{figure}
\label{Fig4}
  \centering
  \includegraphics[scale=0.25]{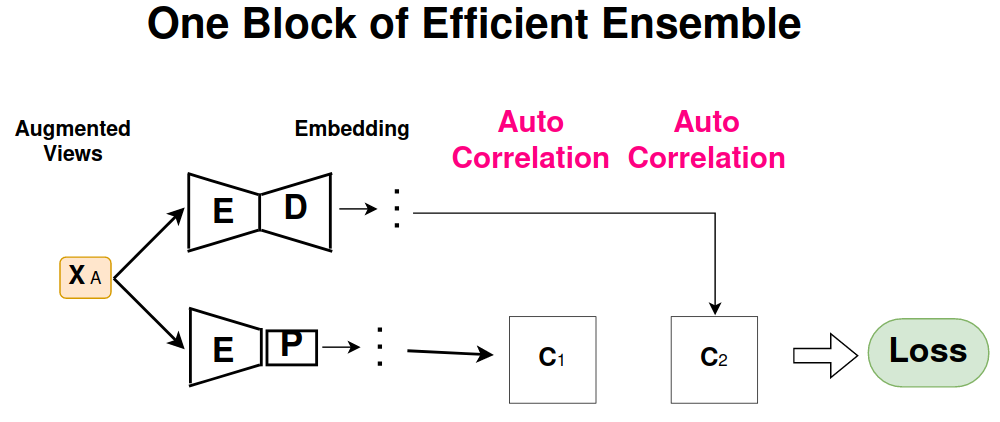}
		\caption{More efficient GUESS framework. For more efficiency, the building block of the ensemble is simplified as shown. Even using this design with one autoencoder (GUESS-1-E) our method outperform former baselines with no extra computational overhead (similar computational complexity to that of BarlowTwins).} 
\end{figure}
\section{Experiments and Results}
We performed  experiments on six benchmark datasets to evaluate the performance of our methods, and to compare with state-of-the-art (SOTA) baselines under identical settings. Later in an ablation study, we also evaluate the impact of both contributions of this work in isolation.

\indent
\textbf{Baselines:} GUESS will be contrasted against six baselines, including SimCLR \cite{chen2020simple}, SwAV \cite{caron2020unsupervised}, BYOL \cite{grill2020bootstrap}, SimSiam \cite{chen2021exploring}, Whitening-MSE ($d=4$) \cite{ermolov2021whitening} and Barlow Twins \cite{zbontar2021barlow}.
SimCLR is a contrastive baseline in which formulation in the original paper \cite{chen2020simple} with $\tau=0.1$ is used. SwAV is clustering based method, BYOL and SimSiam are pioneering non-contrastive baselines. Finally, Whitening-MSE and Barlow Twins are recent baselines, 
that use whitening with non-contrastive loss. 
\subsection{Experimental setting}
\textbf{Datasets and augmentation protocols:} Six benchmark datasets are used for the evaluation of the work at different scales, namely, CIFAR10, and CIFAR100 \cite{krizhevsky2009learning}, Tiny ImageNet \cite{le2015tiny}, ImageNet \cite{le2015tiny}, VOC0712 (detection), and COCO (segmentation).
The standard augmentation for training data is performed under a $p$ random distribution of augmentation including random crop, random aspect ratio adjustment, horizontal mirroring and color jittering. More specifically, for CIFAR10/100, ImageNet Tiny ImageNet, VOC0712, and COCO, following Chen at al \cite{chen2020simple}, a random crop of size $0.2-1$ of the original image size is performed. Moreover, the aspect ratio of the augmented view has a random distribution covering $0.75$ to $1.\Bar{3}$ of the original aspect ratio. Following the configuration in \cite{chen2020simple}, color jittering with a probability ratio of $8/1$ for configuration $(0.4, 0.4, 0.4, 0.1)$ to grayscaling is performed. Finally horizontal mirroring is performed with a probability mean of $0.5$.   
As shown in Fig. 1
depending on the size of the ensemble $M$, the augmentation with distribution $p$, outputs $2M$ batches of augmented examples per one batch of original images.

In case of ensemble, each block of ensemble is trained only on two batches of the augmented instances and not all of them. That is, each pair of $2M$ batches is randomly allocated to a certain block of the ensemble. 

Similarly, in case of efficient ensemble, all 2$M$ pairs of views are allocated in a way that  each block of ensemble is trained only on two batches of the augmented instances and not all of them. \\
\indent
\textbf{Architectures:} Following the details in \cite{ermolov2021whitening}, we use two architectures for the network encoders comparing all baselines. In all experiments on CIFAR10 and CIFAR100, except for the transfer learning, the encoder is ResNet18, while for Tiny ImageNet, ImageNet, VOC0712, and COCO, ResNet50 \cite{he2016deep} is adopted as the encoder. Specifically, similar to \cite{zbontar2021barlow}, we replaced the last layer of both ResNet18 and ResNet50 with a projector head consisting of three consecutive linear layers of individual size of 1024 and 2048 respectively, with the first and second layers each followed by batch norm and ReLU. Note that the output of the third layer of the projector head is used to calculate the cross-correlation (or auto-correlation in case of efficient ensemble). 

For encoder part of the autoencoder in our method, we used the similar architecture to that of the network encoder except that the three-layer projector head is replaced with one fully connected layer of the same size as the autoencoder's embedding space. The autoencoder is symmetric and the architecture of the decoder is a mirror image or a reverse of the encoder's architecture. Each autoencoder is pre-trained for 250 epochs before placing in the pipeline, unless otherwise specified.\\
\indent

\textbf{Implementation details:} The optimisation process for experiments with all datasets was performed using the Adam optimizer \cite{kingma2014adam}. For transfer learning with ResNet50, Tiny ImageNet images (pre-training time) and CIFAR10/100 images (evaluation time) resized to match the input size.
For CIFAR10 and CIFAR100 the training starts with 20 epochs of warm-up with learning rate of $0.15$, and continue with a learning rate of $0.001$ for a total of 1000 epochs with a weight decay  of $10^{-6}$. For ImageNet and Tiny ImageNet the warm-up learning rate is $0.2$, which changes to $0.001$ after 10 epochs, for a total of 1000 epochs with the same decay. Following \cite{ermolov2021whitening}, the batch size during training process for all datasets is set to 1024. For transfer learning on VOC0712 (detection task) and COCO (segmentation task) we follow the settings in \cite{zbontar2021barlow}.
In the ablation study we investigate the case in which autoencoders do not receive any pre-training. Best values for  hyperparameters $\alpha$ and $\beta$ are found to be $0.2$ and $0.01$, respectively. The weight decay is set to $10^{-6}$. 
Unless otherwise stated, for both ensemble and approximate ensemble architectures, in all experiments (including ablation study), all autoencoders are pre-trained for 250 epochs on their specifically allocated data samples, with a learning rate of $0.1$ and $0.001$ for first 10 epochs and remaining epochs having a weight decay, respectively.
\subsection{Initial experiments on pseudo-whitening}
As promised in Section \ref{sec3into}, we present a experimental baseline to contrast pseudo-whitening with whitening. Specifically we present the results of the Barlow-Twins in three different setting. First Barlow-Twins with strict whitening constraint (off-diagonal elements set to zero), then the results of whitening with a relaxed constraint. Finally the results of regularizing the whitening by adding some noise, a predefined matrix with Gaussian distributed off-diagonal elements. Assessing Barlow-Twins with ResNet50 as the building architecture, trained on Tiny ImageNet for 500 epochs (similar settings to those of ablation study), the three cases are evaluated in relation to $\lambda$ in its loss function:\\
\begin{equation}
    \label{eq6}
    \footnotesize
    \begin{split}
     & \quad \mathcal{L}_{BT} \triangleq \sum_{i}(1-C_{ii})^2 + \lambda\sum_{i}\sum_{j\neq i}(C_{ij})^2,\\ & C_{ij}\triangleq \frac{\sum_m z_{m,i}^{A} z_{m,j}^{B}}{\sqrt{\sum_m (z_{m,i}^{A})^2}\sqrt{\sum_m (z_{m,j}^{B})^2}}
     \end{split}
\end{equation}

\textbf{A) $\lambda = 0.005$:} This case is same as the Barlow-Twins setting, as is considered as the baseline for the two following cases. The top-1 accuracy result for Tiny ImageNet with above mentioned settings, is $50.04\%$.

\textbf{B) $\lambda =0.01, 0.1:$} This is the case in which $\lambda$ grows to 2 and 20 times its value for the original Barlow-Twins setting respectively. In essence, here the decorrelation process (whitening) is more strict than the original case in Barlow-Twins paper as bigger $\lambda$ weighs more on the importance of second term of the loss function. The top-1 accuracy result attained with this setting are $49.78$ and $45.48\%$ respectively. Compared to the original case (above), we observe that more strict decorrelation does not necessarily improve the performance, as it is theoretically expected to lead to better results with more redundancy reduction.

\textbf{C) $\lambda = 0.005$} for the following loss function:
\begin{equation}
    \label{eq6}
    \footnotesize
    \begin{split}
     &\quad \mathcal{L}_{BT} \triangleq \sum_{i}(1-C_{ii})^2 + \lambda\sum_{i}\sum_{j\neq i}(C_{ij}-G_{ij})^2, \\
     & C_{ij}\triangleq \frac{\sum_m z_{m,i}^{A} z_{m,j}^{B}}{\sqrt{\sum_m (z_{m,i}^{A})^2}\sqrt{\sum_m (z_{m,j}^{B})^2}}\\
     \end{split}
\end{equation}

where $G$ is an empirical cross-correlation matrix computed from a pair of autoencoders latent space. Specifically the same set of distorted versions feeding to the Barlow Twins, is firstly fed to the autoencoders, to compute the cross-correlation matrix after training for some 500 epochs. Then Barlow-Twins is trained with the new loss function, setting diagonal elements to 1 and off-diagonal elements to their corresponding elements in $G$, which are normally close to zero. The top-1 accuracy here is $50.51\%$. The same experiment with $G$ being a square symmetric matrix its upper triangle elements sampled from a multivariate Gaussian with $\mu=[0,0]$ and $\Sigma=[1,0;0,1]$, results in a top-1 accuracy of $50.11\%$. Regarding the relaxed redundancy reduction constraint, it is inferred that with both $G$s, performing pseudo-whitening by adding some uncertainty to the loss function either brings results on par with the original setting, or improves upon it. We see this as a \textbf{regularization method to the rigorous whitening}, which allows regularized redundancy reduction.
\begin{table*}
  \caption{\footnotesize
  Top-1 classification accuracy on CIFAR10/100, Tiny ImageNet and ImageNet under supervised linear evaluation and KNN with $k=5$ for both ensemble and efficient ensemble models. Even with one block, GUESS-1, or GUESS-1-E, our method outperforms the state-of-art on three dataset including ImageNet, under linear evaluation. Further, we can observe that compared with the improvements offered by B-Twins and W-MSE4 over their former baselines, GUESS offers more relative improvements on CIFAR10, CIFAR100, and ImageNet. We also presented ImageNet transfer learning results with VOC0712 and COCO for detection and segmentation tasks, showing the effectiveness of the building blocks, GUESS-1 and GUESS-1-E.}
  \label{table1}
  \centering
  \scriptsize
  \begin{tabular}{p{1.99cm} p{.75cm} p{.75cm}|p{.75cm} p{.75cm} |p{.75cm} p{.75cm} |p{.75cm} p{.75cm}|p{1.2cm}p{1.2cm}|p{1.2cm}}
    \toprule
    Framework   & \multicolumn{2}{c}{CIFAR10}     & \multicolumn{2}{c}{CIFAR100} & \multicolumn{2}{c}{Tiny ImageNet} & \multicolumn{2}{c}{ ImageNet} & \multicolumn{2}{c}{ VOC0712 (Det.)}  & \multicolumn{1}{c}{ COCO (Seg.) }\\
    \cmidrule(r){2-12}
        & Linear     & KNN &  Linear & KNN & Linear & KNN & Linear & KNN &  A100 &  A50      & A50  \\
    \cmidrule(r){1-12}
    SimCLR   & 91.87 & 88.42 & 66.83 & 56.56 & 49.14 & 35.93 &  69.5 & 53.9 &  56.3 & 82.3 & 55.5   \\
    BYOL   & 91.73 &  89.45  & 66.60  & 56.82  &  51.40  & 36.63 & 74.4 & 54.7 & \textbf{ 57.1 } & \textbf{82.8} &   55\textbf{}.8   \\
    SwAV   & 92.21 &  88.54  &  67.55 & 56.31 &  51.59 &  36.79 & \textbf{75.3} & 53.1 &  56.1   & 82.6  &  55.2    \\
    SimSiam  & 92.11     & 89.31 &66.24  & 56.42& \textbf{51.66} &  \textbf{37.2} & 71.4 & 54.9  & 57  &  82.4 &   \textbf{56}      \\
    W-MSE4   & 91.99 &  \textbf{89.87}  & \textbf{67.64}  & 56.45 &  49.84 &  35.96 & 73.4 & 55.1 &  56.1 & 81.3 &  55.8    \\
    B-Twins   & \textbf{92.33}     & 88.96 & 67.44  & \textbf{57.61} & 50.56 &36.64  & 73.6 &\textbf{55.8} & 56.8  & 82.6   &  \textbf{56} \\
    \hline
    GUESS-1 (ours)   & \textbf{92.78 }    & 89.24 &  \textbf{67.67} & 56.91& 51.43&  36.12 & \textbf{75.6} & 55.3 &  \textbf{57.4} & \textbf{83.7}  &  \textbf{55.9}  \\
    
     GUESS-3 (ours)   & \textbf{93.56}     & \textbf{90.21} &  \textbf{68.41} & 57.54& \textbf{52.14} &  37.11 & \textbf{76.2} & \textbf{55.9} &  - &  - &  -  \\

     GUESS-5 (ours) & \textbf{93.85}     & \textbf{90.67} &  \textbf{68.63} & \textbf{57.42}& \textbf{52.33}&  \textbf{37.63} & \textbf{76.6} & \textbf{56.1} & -  &  - &  -  \\
     \hline
     GUESS-1-E (ours)  & \textbf{92.69 }    & 89.00 &  {67.53} & 56.60 & 51.35&  35.02 & \textbf{75.4} & 54.9 &  \textbf{57.4} &  \textbf{83.4} &   55.7 \\
     GUESS-3-E (ours)   & \textbf{93.48}     & \textbf{90.04} &  \textbf{68.29} & 57.38& \textbf{51.93} &  37.00 & \textbf{76.1} & \textbf{55.5} &  - &  - &  -  \\
    \bottomrule
  \end{tabular}
\end{table*}
\subsection{Evaluation}
After self-supervised feature learning, the standard protocol to evaluate the  SSL approach is to use the fixed pre-trained ConvNet followed by a linear classifier in a supervised manner. Specifically here one encoder per block of the model will be used in another architecture in which, the encoder is followed by a fully connected layer and a softmax.

Following the details of evaluation in \cite{ermolov2021whitening}, we train this architecture ($m$ encoder in case of ensemble, one encoder in case of approximate ensemble) for 500 epochs on the original samples. For ensemble representation depending on the ensemble size $M$, for $M\neq 1$ we have a set of ConvNets followed by the classifier to evaluate the performance. Therefore, after training them, over the test time the accuracy is calculated based on the majority vote of the classifiers, recalling that the ConvNet (encoder part of encoder-projector) of each of the models (with ConvNet followed by classifier) is self-trained on different versions of distorted samples, whereas all the classifiers are trained and then tested over the original samples. In case of inconclusive majority vote, (i.e., the test label of all blocks of the ensemble ($M$) are $M$ different class labels), the label with the highest top-5 rank in other blocks is chosen as the label for the test sample. 
In case of efficient ensemble, the evaluation is performed in similar manner.

 Weight decay for supervised training over evaluation stage is $10^{-6}$ and learning rate starts with $10^{-3}$ and exponentially decays to $10^{-6}$. The Adam optimizer is used for training over the evaluation process. As presented in Table \ref{table1}, and similar to \cite{ermolov2021whitening}, we also examine the classification accuracy using $k$-nearest neighbours (KNN) with $k=5$ which is essentially immediately performed on top of the ConvNets with frozen weights, without further training.
We also pre-train GUESS built of standard ResNet50 on Tiny ImageNet and ImageNet and perform transfer learning with CIFAR10/100 (Table \ref{table2}) for classification task, and transfer learning with VOC0712 and COCO for detection and segmentation tasks respectively. Standard transfer learning evaluation, similar to the linear evaluation, consists of supervised training with the same setting over the evaluation data, expect that the evaluation data (CIFAR10/100, VOC0712, and COCO) here is other than pre-training data.

\subsection{Results}
\subsubsection{Linear evaluation:} 
Table \ref{table1} represents the results of linear evaluation for six former baselines as well as our proposed GUESS with both ensemble and efficient ensemble. Only the results for W-MSE4 , SimCLR, and BYOL on CIFAR10/100 are reported from \cite{ermolov2021whitening}. 
Thanks to {SOLO Learn \cite{da2022solo}, all other results are reproduced along with our GUESS-1, GUESS-3 and GUESS-5 corresponding to ensemble experiments as well as GUESS-1-E, GUESS-3-E, and GUESS-5-E regarding the efficient ensemble experiments. 
SimCLR (contrastive) and SwAV (clustering based with multi-crop view) approaches still remain competitive with non-contrastive approaches; though B-Twins, W-MSE and GUESS perform slightly better except for ImageNet. SimSiam and to a lower degree BYOL as pioneer negative pair-free baselines,  remain very competitive with SwAV, W-MSE4, B-Twins and GUESS. GUESS-1 is essentially a one-block ensemble, which performs consistently better than B-Twins. Clustering based SwAV is very robust; note that two most recent baselines, B-Twin and W-MSE4, on CIFAR10/100 performed on par with former baseline SwAV, and on Tiny ImageNet and ImageNet slightly worse than SwAV.
\\
\indent 
In case of one-block ensemble, GUESS-1 and GUESS-1-E (with only one block ensemble and one block efficient ensemble) consistently improved over all baselines for CIFAR10, CIFAR100, and ImageNet under linear evaluation showing the effectiveness of our loss function. In case of Tiny ImageNet, GUESS-1 and GUESS-1-E remain very competitive with the state-of-the-art. In case of KNN evaluation, GUESS-1 and GUESS-1-E were also very competitive with the state-of-the-art.
Under both linear and KNN evaluation, with larger ensembles, three or five-block ensembles, the performance of GUESS in both settings, ensemble and efficient ensemble, notably outperform the state-of-the-art, where the performance jumps with $M=3$. 
As presented, even using one block, our approach outperformed all former baselines under linear evaluation.

\subsubsection{Transfer learning:} 
Table 1 and Table 2 present the the results for transfer learning on VOC0712, COCO, and CIFAR10/100 in which the pre-training was performed on ImageNet (for VOC0712, COCO, and CIFAR10/100 datasets) and Tiny ImageNet (for CIFAR10/100) and the evaluation on corresponding datasets. As presented, when pre-training is performed on ImageNet, our approach outperforms all former baselines, even using one block (GUESS-1 or GUESS-1-E), let alone using three or five blocks. Under pre-training on Tiny ImageNet, GUESS-1 remains very competitive while GUESS-3, and GUESS-5 outperform former approaches on CIFAR100. In case of CIFAR10, using pre-training on Tiny ImageNet, GUESS-5 outperforms the state-of-the-art.  Transfer learning results with VOC0712 (detection task) using both GUERSS-1 and GUESS-1-E outperforms former baselines, while in case of COCO (segmentation tasks) our framework remains very competitive with the SOTA.

\section{Ablation study}
We perform ablation study on loss function using Tiny ImageNet, altering the 
parameters of our loss function. Similar to the main experiments, we use a standard ResNet50 pre-trained for 500 epochs instead of 1000 epochs 
to perform a linear classification on the same dataset using \textbf{GUESS-1} (only one block). 
The autoencoders were similarly pre-trained for 250 epochs, prior to the 500 epochs of training the whole block unless otherwise specified.

\textbf{{1) Loss function}:}
The top-1 accuracy attained using the original loss function is $50.91\%$. Then we drop the autoencoders from the architecture as well as loss function (technically returning to Barlow Twins' loss function), which results in a top-1 accuracy of $50.12\%$ (this compares well with Barlow-Twins' result of $50.56\%$ with 1000 epochs).  Next, we perform the same experiment except that autoencoders did not receive a 250 epochs of pre-training before the general training process. The top-1 accuracy after 500 epochs of training with $\beta=0.01$ and $\beta=0.001$ is respectively $48.16\%$ and $49.87\%$. Observing the accuracy vs $\#$ of epochs, it seems that the main effect is that the convergence speed slows down without pre-training the autoencoders.  We also assessed the case of \textbf{no pre-training for autoencoders with \textbf{ImageNet} }, which is presented in Table 3.

We replaced the autoencoder with a variational autoencoder (VAE), with the same size of latent vector; which resulted in notable top-1 accuracy degradation, $49.03\%$. We downsize the latent vector of the autoencoder as well as doing proper size adjustment of the projector layers (down to 1024 from 2048), now the accuracy upgrade to $49.93\%$. We suspect that even though as our framework also concede the claim made by \cite{zbontar2021barlow}, that Barlow Twins notably benefits from higher dimensionality of embedding space, in case of VAE, higher dimensionality in VAE latent space leads to performance degradation in our framework probably due to the additive noise  to the latent space.

\begin{table}
  \caption{\footnotesize
  Top-1 transfer learning classification accuracy for CIFAR10/100 evaluated using ResNet50 pre-trained on Tiny ImageNet and ImageNet. 
}
  \label{table2}
  \centering
  \scriptsize
  \begin{tabular}{p{2.cm} p{1.cm} p{1.cm}|p{1.cm} p{1.cm}}
    \toprule
    Framework   & \multicolumn{2}{c}{CIFAR10}     & \multicolumn{2}{c}{CIFAR100} \\
    \cmidrule(r){2-5}
     & Tiny      &          & Tiny     &  \\
     & ImageNet  & ImageNet & ImageNet & ImageNet \\            
    \cmidrule(r){2-5}
     SimCLR  &   87.58  &  90.21 & 67.27  & 76.21   \\
      BYOL  & \textbf{93.17}    & 91.49   &  68.16 &  78.74   \\
       SwAV  &  92.1  & 94.43 &  68.11 &  \textbf{81.24}   \\
       SimSiam  &  91.63   &  92.91 &  67.99 &  78.60   \\
    W-MSE4  & 91.16  & 95.13   &  \textbf{68.25} & 79.1  \\
   B-Twins  &  91.77   & \textbf{95.33}  &  67.39 &  80.45  \\
\hline
    GUESS-1 (ours)  &   91.85   & \textbf{95.85}  & 67.51  & \textbf{81.66} \\
  
     GUESS-3 (ours)   &  92.66    &  \textbf{96.93} & \textbf{68.93}  & \textbf{81.91}\\
   
     GUESS-5 (ours)  & \textbf{92.93}  & \textbf{97.10} & \textbf{69.19} & \textbf{82.13}  \\
     \hline
       GUESS-1-E (ours)  &   91.82   & \textbf{95.73}  & 67.41  & \textbf{81.51} \\
    \bottomrule
  \end{tabular}
\end{table}
\textbf{2) Ensemble vs efficient ensemble:}
As presented in Tables 1, and 2, the results with efficient ensemble model interestingly remain either on par or very competitive with the results of the corresponding ensemble model. Specifically, in case of linear evaluation on CIFAR10, there is only $0.09\%$ and $0.08\%$ top-1 accuracy difference between GUESS-1 and GUESS-1-E as well as between GUESS-3 and GUESS-3-E respectively. In case of dataset at scale, ImageNet, there is $0.2\%$ and $0.1\%$ top-1 accuracy difference between ensemble and efficient ensemble with respectively one and three blocks. We 
can observe a similar behaviour in case of transfer learning with CIFAR10/100 as well as VOC0712 and COCO. This shows the effectiveness of the modification trick presented in Section 3.3 to reduce the 
complexity while keeping the desired performance. 
Thus, efficient ensemble could offer desirable results, on par with ensemble, with only half the computational complexity.  

\textbf{3) Robustness and training on shared data:}
We employed another set of augmentation along with standard augmentation, namely heavy augmentation presented by \cite{bai2022directional}, to assess the robustness of the method under the same settings as \cite{bai2022directional}. Accordingly we pre-train and evaluate our framework as well as SimSiam under heavy augmentation protocol (w/RA(2,1)) with Tiny ImageNet. Compared with the original results (SimSiam Top-1 accuracy: 51.66, GUESS-1: 51.43,  GUESS-3: 52.14), both GUESS-1 and GUESS-3 show robustness as the new Top-1 accuracy of SimSiam, GUESS-1, and GUESS-3 are 44.11, 48.03, and 50.11.
We consider a scenario where each block of the ensemble is fed with the same set of distorted views of the original image. We evaluated this idea with GUESS-3 and GUESS-5 applied on Tiny ImageNet, with ResNet50 pre-trained for 500 epochs on all set of augmented views ( 3 sets and 5 sets, respectively). This resulted in top-1 accuracy $51.32\%$, and $51.50\%$, which show slight degradation by $0.19\%$, and $0.28\%$.

\textbf{4) More on beta and convergence speed:}
As shown in Fig. 3, the top-1 accuracy on linear evaluation of CIFAR10 is not very sensitive to $\beta$ (beta) over small variations.
\begin{figure}
\label{Fig_beta}
  \centering
  \includegraphics[scale=.201]{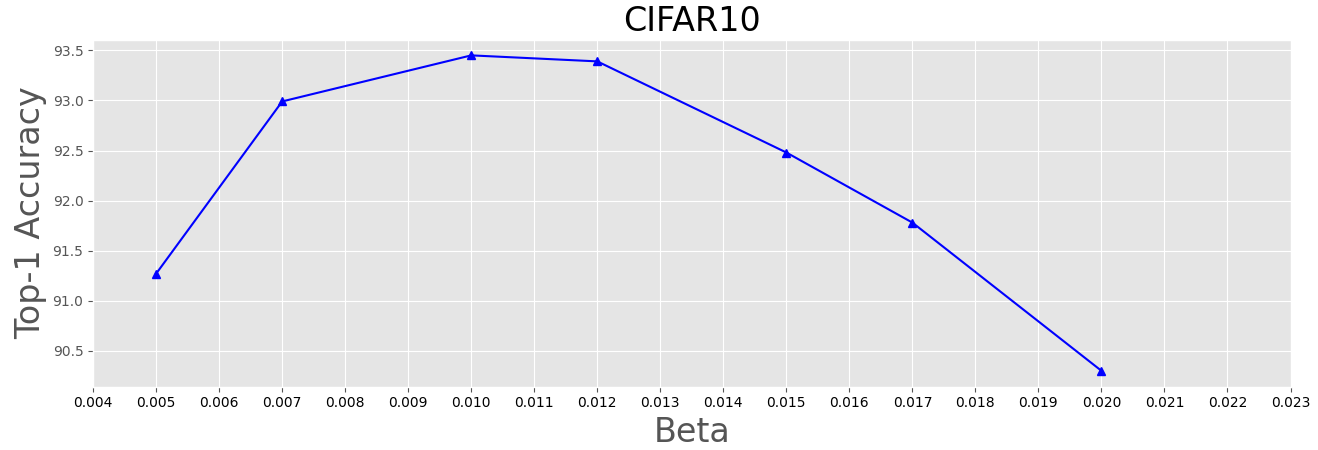}
		\caption{Sensitivity to beta, for CIFAR10, as shown the top-1 linear accuracy ($\%$) is not very sensitive to beta. Best: $\beta=0.01$.}
\end{figure}
Similar to \cite{ermolov2021whitening}, we observe the convergence of accuracy for different methods as the number of epochs increases. In Table 3 the results of different baselines after 200 and 800 epochs of pre-training are evaluated. GUESS-5 is among top three baselines, having faster convergence.
\begin{table}
  \caption{\footnotesize
  Top-1 linear learning classification accuracy convergence for ImageNet using ResNet50. Results for SimSiam and BYOL are from \cite{chen2021exploring}. SwAV is with multi-crop views technique. We also presented results for ImageNet with no pre-training on autoencoder (GUESS-1 (NP)).
}
  \label{table2}
  \centering
  \footnotesize
  \begin{tabular}{p{2.4cm}| p{1.5cm}| p{1.5cm}}
    \toprule
    Framework   &   200 epochs   & 800 epochs \\
    \midrule
 
     SimCLR  &  68.3  & 70.4    \\
      BYOL  &    \textbf{70.6} & \textbf{74.3}  \\
       SwAV  &  70.1  & \textbf{74.8 }   \\
       SimSiam  &  70   & 71.3  \\
    W-MSE4  & 70.1 &  73.1   \\
   B-Twins  &  \textbf{71}   &  {73.3}   \\
 \hline
     GUESS-5 (ours)  & \textbf{71.4}  & \textbf{76.1}  \\
      \hline
       \hline
     GUESS-1 (NP)  & \textbf{70.6}  & \textbf{74.9}  \\
    \bottomrule
  \end{tabular}
\end{table}

\section{Conclusion}

We presented GUESS, a new SSL framework based on pseudo-whitening composed of a module for controlled uncertainty injection, a new loss function, and architecture to enable a data-dependent invariance enforcement to the augmentation. We performed detailed 
experiments on six benchmark datasets under different settings, and also ablation study to analyze GUESS and evaluate its effectiveness. Comparative analysis with the sate-of-the-art shows that results from GUESS, even using one block set a new baseline on linear and transfer learning evaluation. Using lager ensembles further improved the accuracy. 
Finally, we presented 
a trick to reduce computational complexity by half while keeping the desired performance, which also shows that GUESS-1-E (with no extra computational overhead) still outperforms former approaches. 

\bibliographystyle{IEEEtran}
\bibliography{ref} 

\clearpage

\appendices

\section{Efficient ensemble with Auto-correlation}
As discussed in the paper, we propose \textbf{efficient ensemble}, in which we reduce the computational complexity by half via a simpler architectural design and loss function, substituting the cross-correlation with auto-correlation. Corresponding loss function is as follows:

\begin{equation}
    \label{eq1}
    \footnotesize
     \quad \mathcal{L}_{w'} \triangleq \sum_{i}(1-C'_{ii})^2 + \beta\sum_{i}\sum_{j\neq i}(C'_{ij}-C"_{1,ij})^2 ; \quad \mathcal{L}_r =\mathcal{L}_{r_1} + \mathcal{L}_{r_{2}}
\end{equation}
where we have:
\begin{equation}
    C'_{ij} \triangleq \frac{\sum_{m'} z_{m',i}z_{m',j}}{\sqrt{\sum_m' (z_{m',i})^2} \sqrt{\sum_m' (z_{m',j})^2}}
\end{equation}
 where $z$ is the normalized output of projector head for one view, $x_1$, and $m'$ is  the batch size (note that similar to the  original framework, here for each sample we fed two views to the network).  Finally the elements of the matrix $C"$, auto-correlation matrix, is also computed from the latent space of one autoencoder similar to the equation for the elements of $C'$. 

\section{Preliminary}\subsection{Loss functions}
Below, we briefly discuss several objective functions, including triplet loss, typical contrastive loss, and some non-contrastive loss functions. 

\textbf{Triplet loss:} As a discriminative loss, triplet loss was used in work such as \cite{misra2016shuffle,wang2015unsupervised}, which given three latent spaces $z_j$, $z_j$ and $z_k$, this loss implicitly aims for minimizing the distance between positive pairs ($z_i$ and $z_j$); and maximizing the distance between negative pairs ($z_i$ and $z_k$) as follows:
\begin{equation}
    \label{eq2}
    \mathcal{L}_{\bigtriangleup}=\max(0, z_i^T z_j- z_i^T z_k+m),
\end{equation}
where $m$ is a the margin hyperparameter.  A generalization of triplet loss for joint comparison among more than one negative example, named as multi-class N-pair, is also introduced by \cite{sohn2016improved}
\begin{equation}
    \label{eq1}
  \mathcal{L}_{z_i,z_J} = \log\left( 1+\sum_{k=1,k\neq i}^{2N}\exp(z_i z_k - z_i z_j)\right)
\end{equation}
\textbf{Contrastive loss:} Even though the most applicable discriminative loss in representation learning up until recently \cite{oord2018representation,tian2020contrastive,he2020momentum,chen2020simple,bachman2019learning}, is contrastive loss, this is a very demanding loss in terms of number of negative examples required to be contrasted against each positive example. Updated version of this loss are also still demanding either computationally or in terms of negative batch size. \cite{wang2015unsupervised} reformulates the basic contrastive loss in representation learning with $N-1$ negative examples and $\tau$ is a temperature hyperparameter as follows:
\begin{equation}
    \label{eq3}
    \mathcal{L}_{Contrastive} = -\log \frac{\exp (z_i^T z_j/\tau)}{\sum_{n=1, n\neq i}^{N}\exp (z_i^T z_k/\tau)}.
\end{equation}

\textbf{Non-contrastive loss functions:} Almost all self-supervised learning frameworks building on contrastive loss, require negative pairs in their strategy to explore the representation space. Moreover, unlike its popularity, contrastive loss and its variants consistently tend to tightly tie the performance to the number of negative instances per batch.To address this issue and other issues including representation collapse to trivial solution, pioneer work on non-contrsative loss (non-contrastive approach only relies on positive sample pairs) including BYOL \cite{grill2020bootstrap} and later SimSiam \cite{chen2021exploring} devised a type of loss which is needless of negative instances while avoiding representation collapse. In particular BYOL shows that negative instances
are dispensable and the framework surprisingly avoids the representation collapse caused by utilizing large batch of negative examples.
This new line of approaches which relies only on positive pairs in return of architecture update (adding a predictor block) as well as training protocol (stop-gradient policy), substantially outperform contrastive approaches. Tian et al. \cite{tian2021understanding} fundamentally investigate these works to realize the cause of eliminating representation collapse, which reveal a few protocols playing a central role on navigating towards non-trivial representations. 

Along with this line, most recently two approaches know as whitening-MSE and Barlow Twins \cite{ermolov2021whitening,zbontar2021barlow} primarily based on whitening the embedding or batches of embedding space defined new baselines. Whitening-MSE also called hard whitening \cite{ermolov2021whitening} applies Cholesky decomposition to perform whitening over embeddings of a pair of networks, followed by a cosine similarity between the output of operation from each  of networks as follows:
\begin{equation}
    \label{eq4}
    \min_{\theta} \mathbb{E}[2-2\frac{\langle z_i,z_j \rangle}{\lVert  z_i  \rVert_2 . \lVert  z_j \rVert_2}],\quad s.t.\:cov(z_i,z_i)=cov(z_j,z_j)=I.
\end{equation}
{which ends up using MSE}

Barlow Twins \cite{zbontar2021barlow} also called soft whitening performs whitening over the square matrix cross-correlation of twin networks outputs, $C$, which has relatively simpler loss function as follows:
\begin{equation}
    \label{eq6}
    \begin{split}
     &\quad \mathcal{L}_{BT} \triangleq \sum_{i}(1-C_{ii})^2 + \lambda\sum_{i}\sum_{j\neq i}(C_{ij})^2, \\
     & C_{ij}\triangleq \frac{\sum_m z_{m,i}^{A} z_{m,j}^{B}}{\sqrt{\sum_m (z_{m,i}^{A})^2}\sqrt{\sum_m (z_{m,j}^{B})^2}}\\
     \end{split}
\end{equation}
where $\lambda>0$ is a trade-off constant with typical value of $10^{-2}$, $m$ goes over batch samples, $i, j$ are indices bounded by the dimension of the outputs of the networks and $-1\leq C_{ij}\leq1$.

One last point worth mentioning, is that a very recent analytical work on loss functions by Balestriero and LeCun \cite{balestriero2022contrastive} , suggests that non-contrastive loss functions are generally more preferable due to better error bound on downstream tasks.

Aside from above mentioned methods which directly involve features, a different set of approaches based on clustering \cite{caron2020unsupervised,caron2018deep,caron2018deep,caron2021emerging} primarily using cross-entropy loss in geometrical setting also have emerged which is not the focus of this work.

\subsection{Uncertainty and SSL:} 
Some experts including Lecunn \cite{Lexpodcast,zbontar2021barlow} suggest that uncertainty modeling in deep learning, and specifically in self-supervised learning is an under-explored perspective which potentially could bring significant improvement in next decade. That said, model uncertainty in self-supervised learning in not really explored to our best knowledge; except notably a few recent work such as \cite{hendrycks2019using,poggi2020uncertainty,liu2019exploiting}; which are more concerned with impact of self-supervised learning on model uncertainty estimation as well as robustness improvement. For instance \cite{hendrycks2019using} specifically weigh on the other beneficial aspect of self-supervised learning to enhance the performance on downstream task evaluation. Accordingly they leverage self supervision to improve model uncertainty and robustness such as handling adversarial instance and annotation corruption. Another work, \cite{poggi2020uncertainty} poses the importance of accuracy in depth estimation and proposes uncertainty modeling to make the depth estimation more accurate.
With an emphasis on the concept of SSL in robotics and spatial perception, \cite{nava2021uncertainty} proposes a applying uncertainty to former baselines in order to reduce the state estimate error.

\cite{lienen2021credal} translates the concept of credal sets (set of probability distributions) to SSL in order provide model uncertainty in pseudo-labels in low regime labeled data. this work proposes the use of credal sets to model uncertainty in pseudo-labels and hence reduce calibration errors in SSL approach.\\


\subsection{Clustering approaches}
Aside from mentioned methods in the paper, which mainly involve features in a direct way, there is a different set of approaches based on clustering \cite{bautista2016cliquecnn,caron2020unsupervised,caron2018deep,caron2021emerging,ji2019invariant} which involves sample space. Specifically they primarily use cross-entropy loss in geometrical setting to assign a cluster to samples targeting the semantic classes. In terms of loss functions, for each original sample, a pair of augmented views is generated, which one of them guides the loss function to find a target and the other one aims at predicting the same target. This is generally formulated in a framework designed based on geometrical optimisation.  

An interesting point about the clustering-based approaches is that they are also negative-free similar to non-contrastive approaches discussed  in the paper. However, while they are negative pair-free, it does not guarantee the degenrate solution (representation collapse) and also incur computational overhead  due to clustering process.

One canonical example of these set of methods, which is among the state-of-the-art, is SwAV \cite{caron2020unsupervised} in which multiple postitive is used to accomplish sample to cluster allocation via a cross entropy loss optimisation. [put some theory of SwAV's loss function]

\section{Theory}
\subsection{GUESS and Information Bottleneck:}
Rethinking SSL from the lens of information theory, would bring more insight into theoretical aspects of SSL. Here we take the same approach as Barlow-Twins \cite{zbontar2021barlow} to show how GUESS is an instantiation information bottleneck (IB) principle.
IB \cite{tishby2000information} is tasked to find the best trade-off between accuracy and complexity/compression for a random variable $X$ given a joint distribution $p(X,Y)$. Here a better interpretation of that is to view it as a rate distortion problem, with a distortion function that measures how well $X$ is predicted from a compressed representation $Z_\theta$ compared to its direct prediction from $Y$.
Specifically, IB principle asserts that a SSL objective function learns a representation $Z$ which is invariant to random distortion applied to the sample while variant  to (informative of) the sample distribution. This could be formulated as a trade-off between representing sample distribution and disregarding sample distortion as follows:
\begin{equation}
    \label{eq10}
    \mathbb{IB}_\theta \triangleq I(Z_\theta,Y)-\beta I(Z_\theta,X)
\end{equation}
with some basic identity in information theory this could be reformulated as:
\begin{equation}
    \label{eq20}
    \mathbb{IB}_\theta=H(Z_\theta|X)-\frac{1-\beta}{\beta}(H(Z_\theta)
\end{equation}
However, as exact measurement of the entropy is \textbf{computationally deficient} given the batch size of the data, as a classical assumption we reduce the case by assuming that the ${Z}$ is Gaussian distributed, hence its entropy is logarithm of the determinant of its covariance function. Eventually we have:
\begin{equation}
    \label{eq30}
    \mathbb{IB}_\theta=\mathbb{E}_X \log|\mathcal{C}_{Z_\theta|X} |+ \frac{1-\beta}{\beta}\log |\mathcal{C}_{Z_\theta}|
\end{equation}
In practice, GUESS optimizes a slightly different objective function; in fact similar to \cite{zbontar2021barlow}, rather than optimizing the determinant of the covariance matrices, simply minimization on the Frobenius norm of the cross-correlation matrix is performed due to better results in practice. 
Similar to Barlow-Twins, this affects the off-diagonal terms of the covariance matrix, however unlike Barlow-Twins that encourages them to be as close to 0 as possible, here off-diagonal elements are forced to their corresponding constants (which are generally close to zero) extracted form AE generative process. That said , this surrogate objective, which enforces relaxed-decorrelation of all output units, maintains the same global optimum as \ref{eq30}.

\end{document}